# Memory Capacity of a Random Neural Network

Matt Stowe

Abstract: This paper considers the problem of information capacity of a random neural network. The network is represented by matrices that are square and symmetrical. The matrices have a weight which determines the highest and lowest possible value found in the matrix. The examined matrices are randomly generated and analyzed by a computer program. We find the surprising result that the capacity of the network is a maximum for the binary random neural network and it does not change as the number of quantization levels associated with the weights increases.

**Introduction**

It is well known that the capacity of a Hebbian neural network consisting of $n$ neurons is approximately $0.15n$ [1]. This capacity can be increased to $n$ if delta learning is used [2] and substantially more if non-binary networks are used where the capacity depends on how the data is quantized [3].

Learning in biological networks has several aspects that go beyond the models studied by computer scientists [4]-[10], that have both feedback and feedforward types. Amongst these aspects is the finding that the developing brain is characterized by pruning, that is a loss of connections that starts near the time of birth and is completed by the time of sexual maturation in humans [11],[12]. A decrease in the number of synapses occurs after adolescence and approximately half of the neurons during development do not survive to adulthood. Pruning is believed to represent learning although the processes that underlie it are not well understood.

Here we propose that the original random network may itself fulfill an important function and define memories that represent Jungian archetypes associated with the species [13]. In the Jungian view the *tabula rasa* theory of human psychological development is wrong and the original structure brings with it some information. Here we do not consider the ways the original information is coded but it could be related to the structural organization of the brain which carries within it the evolutionary history of the species. We show that the random binary matrices can store a large number of memories.

**The General Model**

Instead of examining Hebbian systems, we use random matrices that follow the same rule of symmetry as a Hebbian system. The diagonal terms of the matrix must all be 0. A memory is represented by a column vector of size n. Memories are represented by only two values, +1 and -1. In general, the + represents any value greater than or equal to 0, and the - represents any value less than 0. A memory is valid for matrix T (the synaptic interconnection matrix) if the dot product of T and the memory is equal to the memory after applying the + and - representation to the dot product.

In other words, a memory is stored if

$$x^i = \text{sgn}(Tx^i) \tag{1}$$

where the sgn function is 1 if the input is equal or greater than zero and -1, if the input is less than 0. The update process we consider is that same as in the Hopfield model although the memories are not learnt by the Hebbian rule.

In the generator model variant of the Hopfield model, the memories are recalled by the use of the lower triangular matrix B, where T = B + B$^t$. In such a model, the activity starts from one single neuron and then spreads to additional neurons as determined by B [14],[15].

The flow of information in this model is determined by the proximity of the neurons and this flow will be different from information starting at different neurons (Figure 1).

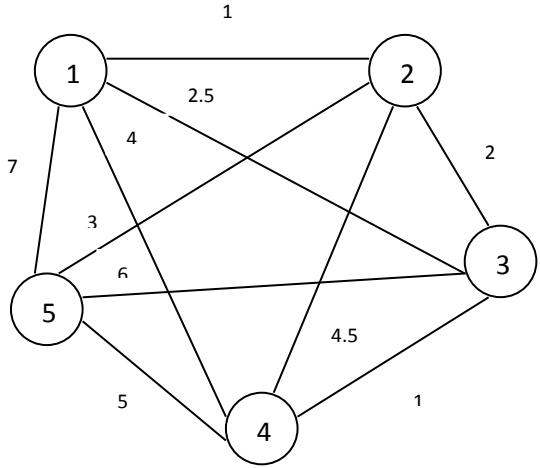

**Figure 1.** Network of five neurons ( from[14])

In Figure 1, the numbers on the links between nodes represent the distance between the nodes and would thus also represent the time taken for a signal to travel over the node. As is clear the information from 1 flows to 2, 3, 4, and 5. But information from 3 will flow to 4, 2, 1, and 5. In other words, the order varies.

**Example System**

Consider the following T matrix which consists of only +1 and -1 entries. In general, the size of such a matrix can be arbitrarily large. Here, we wish to show the major finding of our paper in the context of the example of 10 neurons.

$$\begin{bmatrix} 0 & 1 & 1 & 1 & -1 & 1 & 1 & -1 & -1 & 1 \\ 1 & 0 & -1 & 1 & 1 & -1 & -1 & 1 & 1 & 1 \\ 1 & -1 & 0 & -1 & 1 & -1 & 1 & 1 & 1 & 1 \\ 1 & 1 & -1 & 0 & 1 & -1 & 1 & 1 & 1 & -1 \\ -1 & 1 & 1 & 1 & 0 & 1 & 1 & -1 & -1 & 1 \\ 1 & -1 & -1 & -1 & 1 & 0 & -1 & 1 & -1 & 1 \\ 1 & -1 & 1 & 1 & 1 & -1 & 0 & -1 & -1 & -1 \\ -1 & 1 & 1 & 1 & -1 & 1 & -1 & 0 & -1 & -1 \\ -1 & 1 & 1 & 1 & -1 & -1 & -1 & -1 & 0 & 1 \\ 1 & 1 & 1 & -1 & 1 & 1 & -1 & -1 & 1 & 0 \end{bmatrix} \quad (2)$$

Notice that the weight of this matrix is 1. This means that all values are +1 or -1 except for those along the diagonal of the matrix. The matrix is symmetric and was randomly generated. This matrix had the highest memory capacity out of 100000 tested matrices. It can hold 38 unique memories, not counting the simple inversions.

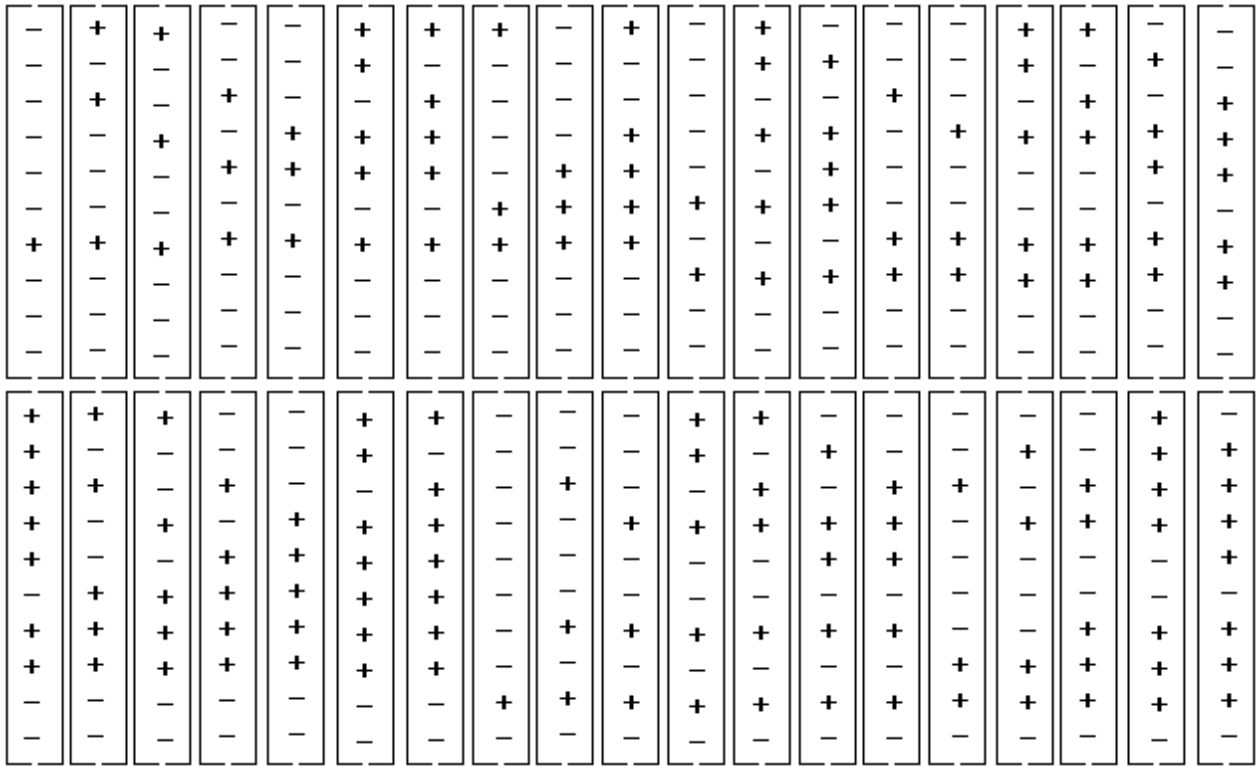

**Figure 1.** The memories stored by the intereconnection matrix of equation 2.

If any of these memories is inverted, then the inversion is also a valid memory for the matrix. This is also generally true of Hebbian networks and the extent it is not true is due to the slight asymmetry in the update process since the 0 value is taken to be 1.

The capacity of this matrix is much higher than that of Hebbian networks or even of a network trained using delta learning.

**Generalization Method**

The above matrix had the highest memory retention of 100000 randomly generated square 10×10 matrices with a weight of 1. Other weights were also tested. Each different weight was tested with 100000 randomly generated square 10×10 matrices, and the matrix with the highest memory capacity was retained along with its memories.

**Results**

The following graph shows the results of the computer program execution. It should be noted that the highest memory capacity is with a matrix weight of 1, and all other weights have mostly the same capacity.

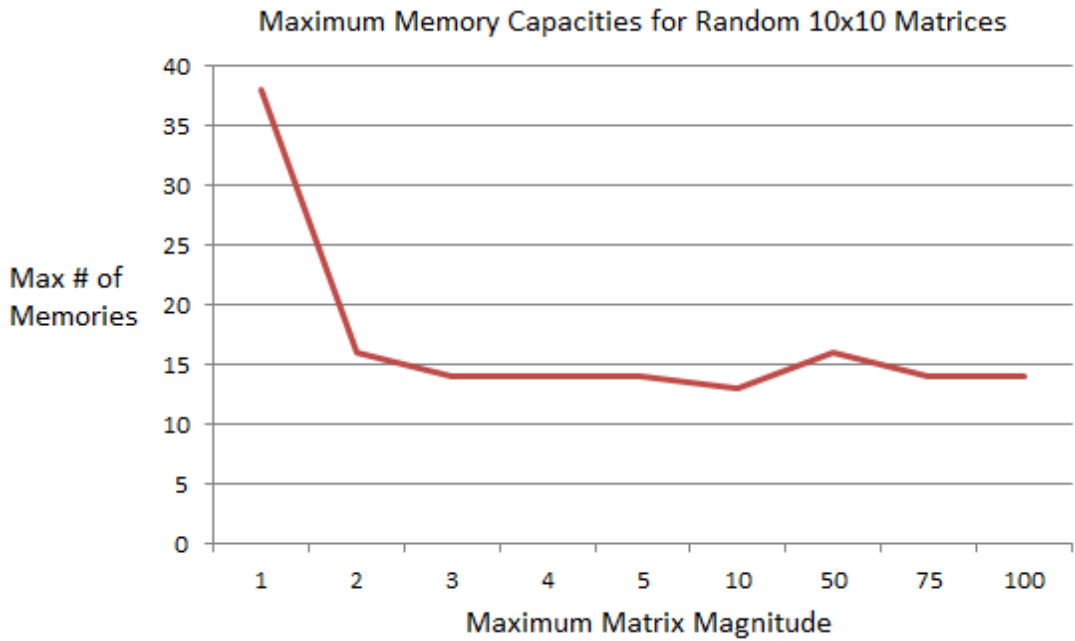

**Figure 2.** The memories stored in relation to the weight of the matrix

The result that the capacity of the network does not change as the weight of the matrix changes is quite surprising and counter-intuitive.

**Conclusion**

The random matrix with a weight of 1 could represent a newborn organism that has not had the chance to learn, and would therefore be unsuited to the Hebbian learning model generally used to model memories. A newborn human has all of its neurons connected, and therefore has a much greater capacity to learn. This could be modeled by a random binary weighted matrix, which has a higher memory capacity than the traditional Hebbian matrix. As the child grows, certain neurons lose their connections, and other neuron connections become stronger. This allows the child to learn, and could be represented by a shift from a random binary matrix to the traditional Hebbian learning model.